\documentclass{article}
\usepackage[english]{babel}
\usepackage[utf8]{inputenc} % allow utf-8 input
\usepackage[T1]{fontenc}    % use 8-bit T1 fonts

\usepackage{csquotes}

\usepackage{arxiv}
\usepackage{url}            % simple URL typesetting
\usepackage{booktabs}       % professional-quality tables
\usepackage{amsfonts}       % blackboard math symbols
\usepackage{nicefrac}       % compact symbols for 1/2, etc.
\usepackage{microtype}      % microtypography

\usepackage{mathtools}

\newcommand*\samethanks[1][\value{footnote}]{\footnotemark[#1]}

\title{Metrics for Multi-Class Classification: an Overview}
\author{
  Margherita Grandini \\
  CRIF S.p.A.\thanks{CRIF S.p.A., via Mario Fantin 1-3, 40131 Bologna (BO), Italy}\\
  \texttt{m.grandini@crif.com}
  \And
  Enrico Bagli \\
  CRIF S.p.A.\samethanks
  \And
  Giorgio Visani \\
  CRIF S.p.A.\samethanks\\
  Department of Computer Science\thanks{Università degli Studi di Bologna, Dipartimento di Ingegneria e Scienze Informatiche, viale Risorgimento 2, 40136 Bologna (BO), Italy},\\
  University of Bologna
}

 \date{\today}

\usepackage{hyperref}
\usepackage{graphicx}
\usepackage{subfig}
\usepackage{caption}
\usepackage{tabularx}
\usepackage{amssymb} 
\usepackage{amsmath}
\usepackage{hyperref}
\hypersetup{
	colorlinks=true,       % false: boxed links; true: colored links
	linkcolor=blue,        % color of internal links
	citecolor=blue,        % color of links to bibliography
	filecolor=magenta,     % color of file links
	urlcolor=blue         
}

\usepackage{biblatex}
\begin{document}

\maketitle

\begin{abstract}
Classification tasks in machine learning involving more than two classes are known by the name of "multi-class classification". Performance indicators are very useful when the aim is to evaluate and compare different classification models or machine learning techniques. Many metrics come in handy to test the ability of a multi-class classifier. Those metrics turn out to be useful at different stage of the development process, e.g. comparing the performance of two different models or analysing the behaviour of the same model by tuning different parameters. In this white paper we review a list of the most promising multi-class metrics, we highlight their advantages and disadvantages and show their possible usages during the development of a classification model.

\end{abstract}

\section{Introduction}

In the vast field of Machine Learning, the general focus is to predict an outcome using the available data. The prediction task is also called "classification problem" when the outcome represents different classes, otherwise is called "regression problem" when the outcome is a numeric measurement.\\
As regards to classification, the most common setting involves only two classes, although there may be more than two. In this last case the issue changes his name and is called "multi-class classification".
\medskip

From an algorithmic standpoint, the prediction task is addressed using the state of the art mathematical techniques. There are many different solutions, however each one shares a common factor: they use available data ($\mathbf{X}$ variables) to obtain the best prediction $\hat{Y}$ of the outcome variable $Y$.\\
In Multi-class classification, we may regard the response variable $Y$ and the prediction $\hat{Y}$ as two discrete random variables: they assume values in $\{ 1, \cdots, K\}$ and each number represents a different class.\\
The algorithm comes up with the probability that a specific unit belongs to one possible class, then a classification rule is employed to assign a single class to each individual.
The rule is generally very simple, the most common rule assigns a unit to the class with the highest probability.\\
A classification model gives us the probability of belonging to a specific class for each possible units. Starting from the probability assigned by the model, in the two-class classification problem a threshold is usually applied to decide which class has to be predicted for each unit. While in the multi-class case, there are various possibilities; among them, the highest probability value and the softmax are the most employed techniques.
\medskip

Performance indicators are very useful when the aim is to evaluate and compare different classification models or machine learning techniques.\medskip

There are many metrics that come in handy to test the ability of any multi-class classifier and they turn out to be useful for: i) comparing the performance of two different models, ii) analysing the behaviour of the same model by tuning different parameters.
\medskip

Many metrics are based on the Confusion Matrix, since it encloses all the relevant information about the algorithm and classification rule performance.

\subsection{Confusion Matrix}

\begin{figure}[ht!]
\centering
\includegraphics[scale=0.7]{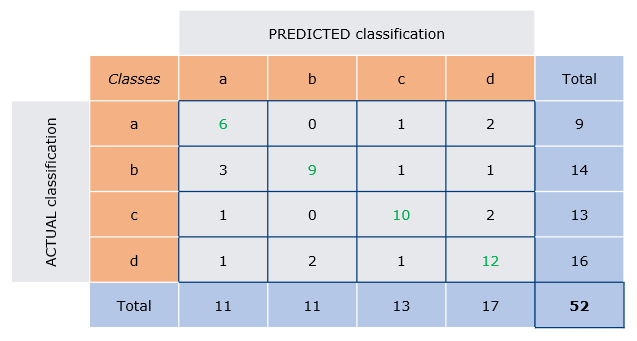}
\caption{Example of confusion matrix}
\label{fig:Multi-class Case}
\end{figure}

The confusion matrix is a cross table that records the number of occurrences between two raters, the true/actual classification and the predicted classification, as shown in Figure \ref{fig:Multi-class Case}. For consistency reasons throughout the paper, the columns stand for model prediction whereas the rows display the true classification.\medskip

The classes are listed in the same order in the rows as in the columns, therefore the correctly classified elements are located on the main diagonal from top left to bottom right and they correspond to the number of times the two raters agree.
\medskip

\medskip

In the following paragraphs, we review two-class classification concepts, which will come in handy later
to understand multi-class concepts.

\subsection{Precision \& Recall}\label{section_prec_rec}

These metrics will act as building blocks for Balanced Accuracy and F1-Score formulas.\\
Starting from a two class confusion matrix: \medskip

 \begin{figure}[ht!]
    \centering
    \includegraphics[scale=0.7]{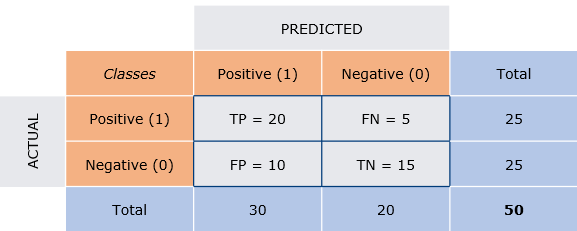}
    \caption{Two-class Confusion Matrix}
    \label{fig:Two classes confusion matrix}
 \end{figure}\medskip

The Precision is the fraction of True Positive elements divided by the total number of positively predicted units (column sum of the predicted positives). In particular, True Positive are the elements that have been labelled as positive by the model and they are actually positive, while False Positive are the elements that have been labelled as positive by the model, but they are actually negative.
\medskip

    \begin{equation}
       Precision = \frac{TP}{TP+FP}
    \end{equation}\medskip

Precision expresses the proportion of units our model says are Positive and they actually Positive. In other words, Precision tells us how much we can trust the model when it predicts an individual as Positive.
\medskip

The Recall is the fraction of True Positive elements divided by the total number of positively classified units (row sum of the actual positives). In particular False Negative are the elements that have been labelled as negative by the model, but they are actually positive.

\begin{equation}
   Recall = \frac{TP}{TP+FN}
\end{equation}

The Recall measures the model’s predictive accuracy for the positive class: intuitively, it measures the ability of the model to find all the Positive units in the dataset.
\medskip

Hereafter, we present different metrics for the multi-class setting, outlining pros and cons, with the aim to provide guidance to make the best choice.

\section{Accuracy}
Accuracy is one of the most popular metrics in multi-class classification and it is directly computed from the confusion matrix.\medskip

Referring to Figure \ref{fig:Two classes confusion matrix}:

\begin{equation}
Accuracy=\frac{TP+TN}{TP+TN+FP+FN}
\end{equation}
\medskip

The formula of the Accuracy considers the sum of True Positive and True Negative elements at the numerator and the sum of all the entries of the confusion matrix at the denominator. True Positives and True Negatives are the elements correctly classified by the model and they are on the main diagonal of the confusion matrix, while the denominator also considers all the elements out of the main diagonal that have been incorrectly classified by the model.\\ 
In simple words, consider to choose a random unit and predict its class, Accuracy is the probability that the model prediction is correct.
\medskip

Referring to Figure \ref{fig:Multi-class Case}:

\begin{equation}
Accuracy=\frac{6+9+10+12}{52}
\end{equation}
\medskip

The same reasoning is also valid for the multi-class case.
\medskip

Accuracy returns an overall measure of how much the model is correctly predicting on the entire set of data. The basic element of the metric are the single individuals in the dataset: each unit has the same weight and they contribute equally to the Accuracy value. \\
When we think about classes instead of individuals, there will be classes with a high number of units and others with just few ones. In this situation, highly populated classes will have higher weight compared to the smallest ones.
\medskip

Therefore, Accuracy is most suited when we just care about single individuals instead of multiple classes. The key question is "Am I interested in a predicting the highest number of individuals in the right class, without caring about class distribution and other indicators?". If the answer is positive, then the Accuracy is the right indicator.  
\medskip

A practical example is represented by imbalanced datasets (when most units are assigned to a single class): Accuracy tends to hide strong classification errors for classes with few units, since those classes are less relevant compared to the biggest ones. \\
Using this metric, it is not possible to identify the classes where the algorithm is working worse.
\medskip

On the other hand, the metric is very intuitive and easy to understand. Both in binary cases and multi-class cases the Accuracy assumes values between $0$ and $1$, while the quantity missing to reach $1$ is called $Misclassification Rate$ \cite{CHOI1986173}.
\medskip

\section{Balanced Accuracy}
Balanced Accuracy is another well-known metric both in binary and in multi-class classification; it is computed starting from the confusion matrix. \medskip

Referring to Figure \ref{fig:Two classes confusion matrix}: 
\begin{equation}\label{bal_acc_two}
Balanced\ Accuracy=\frac{\frac{TP}{Total_{row_1}}+ \frac{TN}{Total_{row_2}}}{2}
\end{equation}\medskip

Referring to Figure \ref{fig:Multi-class Case}:
\begin{equation}\label{bal_acc_multi}
Balanced\ Accuracy=\frac{\frac{6}{9}+ \frac{9}{14} + \frac{10}{13}+ \frac{12}{16}}{4}
\end{equation}\medskip

The formula of the Balanced Accuracy is essentially an average of recalls.
First we evaluate the Recall for each class, then we average the values in order to obtain the Balanced Accuracy score. The value of Recall for each class answers the question "how likely will an individual of that class be classified correctly?". Hence, Balanced Accuracy provides an average measure of this concept, across the different classes.
\medskip

If the dataset is quite balanced, i.e. the classes are almost the same size, Accuracy and Balanced Accuracy tend to converge to the same value.\\
In fact, the main difference between Balanced Accuracy and Accuracy emerges when the initial set of data (i.e. the actual classification) shows an unbalanced distribution for the classes.
\medskip

\begin{figure}[ht!]
    \centering
    \includegraphics[scale=0.7]{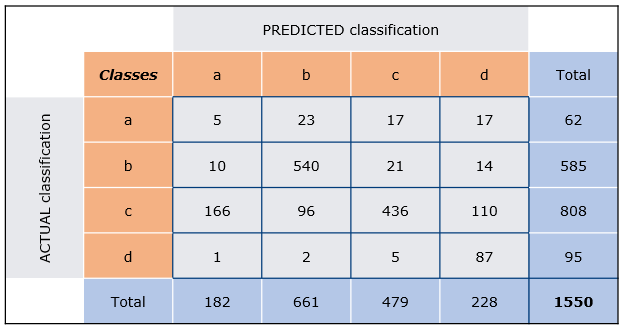}
    \caption{Imbalanced Dataset}
    \label{fig:Imbalanced Dataset}
\end{figure}
\medskip

Figure \ref{fig:Imbalanced Dataset} shows how the actual classification is unbalanced towards classes "b" and "c". For this setting, Accuracy value is 0.689, whereas Balanced Accuracy is 0.615. The difference is mainly due to the weighting that recall applies on each row/actual class. In this way each class has an equal weight in the final calculation of Balanced Accuracy and each class is represented by its recall, regardless of their size. Accuracy instead, mainly depends on the performance that the algorithm achieves on the biggest classes. The performance on the smallest ones is less important, because of their low weight.
\medskip

Summarizing the two main steps of Balanced Accuracy, first we compute a measure of performance (recall) for the algorithm on each class, then we apply the arithmetic mean of these values to find the final Balanced Accuracy score. All in all, Balanced Accuracy consists in the arithmetic mean of the recall of each class, so it is "balanced" because every class has the same weight and the same importance.\\
A consequence is that smaller classes eventually have a more than proportional influence on the formula, although their size is reduced in terms of number of units. This also means that Balanced Accuracy is insensitive to imbalanced class distribution and it gives more weight to the instances coming from minority classes. On the other hand, Accuracy treats all instances alike and usually favours the majority class \cite{5597285}.
\medskip

This may be a perk if interested in having good prediction also for under-represented classes, or a drawback if we care more about good prediction on the entire dataset. \\
The smallest classes when misclassified, are able to drop down the value of Balanced Accuracy, since they have the same importance as largest classes have in the equation. For example, considering class "a" in the Figure \ref{fig:Imbalanced Dataset}, there are 57 misclassified elements and 5 elements which have been rightly predicted, for a total row of 62 elements belonging to the class "a" observing the actual classification. An amount of 57 elements have been assigned to other classes by the model, in fact the recall for this small class is quite low (0.0806).
\medskip

When the class presents a high number of individuals (i.e. class "c"), its bad performance is caught up also by the Accuracy. Instead, when the class has just few individuals (i.e. class "a"), the model's bad performance on this last class cannot be caught up by Accuracy. If we are interested in achieving good predictions (i.e. class "b" and "d") also for rare classes, the information of Balanced Accuracy guarantees to spot possible predictive problems also for the under-represented classes.
\medskip

% \begin{table}[h]
% \begin{tabularx}{\textwidth}{|l|X|}
% \hline
% \textbf{Balanced\ Accuracy = 1}: & The algorithm has perfectly classified all the classes. \\
% \hline
% \textbf{$0<\;$Balanced\ Accuracy$\;<1$}: & There are some elements out of the main diagonal that have been badly predicted by the algorithm. More the value is close to 0, more the elements have been badly classified.\\
% \hline
% \end{tabularx}
% \end{table}
% \medskip

\subsection{Balanced Accuracy Weighted}

The Balanced Accuracy Weighted takes advantage of the Balanced Accuracy formula multiplying each recall by the weight of its class $w_k$, namely the frequency of the class on the entire dataset. We add also the sum of the weights $W$ at the denominator, with respect to the Balanced Accuracy.

\begin{equation}
Balanced\ Accuracy\ Weighted= \frac{\sum_{k=1}^K\frac{TP_k}{Total_{row_k}\cdot w_k}}{{K\cdot W}}
\end{equation}

Referring to Figure \ref{fig:Multi-class Case}:

\begin{equation}
Balanced\ Accuracy\ Weighted=\frac{\frac{6}{9}\cdot w_a+ \frac{9}{14}\cdot w_b + \frac{10}{13}\cdot w_c+ \frac{12}{16}\cdot w_d}{4\cdot W}
\end{equation}

Once recalls have been weighted by the frequency of each class ($w_k$), the average of recall is no longer dirtied by low frequency classes: large classes will have a proportional weight to their size, and small ones will have a resized effect, compared with the Balanced Accuracy formula.
\medskip

Since every recall is weighted by the class frequency of the initial dataset, Balanced Accuracy Weighted could be a good performance indicator when the aim is to train a classification algorithm on a wide number of classes. In fact, this metric allows to keep separate algorithm performances on the different classes, so that we may track down which class causes poor performance. At the same time, it keeps track of the importance of each class thanks to the frequency. This ensures to obtain a reliable value of the overall performance on the dataset: we may interpret this metric as the probability to correctly predict a given unit, even if the formula is slightly different from the Accuracy.

\section{F1-Score}
Also F1-Score assesses classification model's performance starting from the confusion matrix, it aggregates Precision and Recall measures under the concept of harmonic mean.\medskip

    \begin{equation}
     \textrm{F1-Score}=\left( \frac{2}{precision^{-1}+recall^{-1}}\right) = 2\cdot \left( \frac{precision\cdot recall}{precision+recall}\right)
    \end{equation}
\medskip

%According to the theory of measurement the composite measure should satisfy the following 6 definitions:

%Connectedness(two pairs can be ordered) and transitivity(if e1 >= e2 and e2 >= e3 then e1 >= e3)
%Independence: two components contribute their effects independently to the effectiveness.
%Thomsen condition: Given that at a constant recall (precision) we find a difference in effectiveness for two values of precision (recall) then this %difference cannot be removed or reversed by changing the constant value.
%Restricted solvability.
%Each component is essential: Variation in one while leaving the other constant gives a variation in effectiveness.
%Archimedean property for each component. It merely ensures that the intervals on a component are comparable.
%We can then derive and get the function of the effectiveness:
%https://stackoverflow.com/questions/26355942/why-is-the-f-measure-a-harmonic-mean-and-not-an-arithmetic-mean-of-the-precision

The formula of F1-score can be interpreted as a weighted average between Precision and Recall, where F1-score reaches its best value at 1 and worst score at 0. The relative contribution of precision and recall are equal onto the F1-score and the harmonic mean is useful to find the best trade-off between the two quantities \cite{sasaki2007truth}.\medskip

The addends "Precision" and "Recall" could refer both to binary classification and to multi-class classification, as shown in Chapter \ref{section_prec_rec}: in the binary case we only consider the Positive class (therefore the True Negative elements have no importance), while in the multi-class case we consider all the classes one by one and, as a consequence, all the entries of the confusion matrix. \medskip

To give some intuition about the F1-Score behaviour, we review the effect of the harmonic mean on the final score.\\
It has been observed from previous studies that it gives large weight to smaller classes and it mostly rewards models that have similar Precision and Recall values. As an example, we consider Model A with Precision equal to Recall (80\%), and Model B whose precision is 60\% and recall is 100\%. Arithmetically, the mean of the precision and recall is the same for both models, but using the harmonic mean, i.e.  computing the F1-Score, Model A obtains a score of 80\%, while Model B has only a score 75\% \cite{shmueli_2019}.
\medskip

Moreover, Precision and Recall take values in the range [0;1] and when one of them assumes values close to 0, the final F1-Score suffers a huge drop. In fact the harmonic mean tends to give more weight to lower values.

\subsection{F1-Score Binary case}
Referring to confusion matrix in Figure \ref{fig:Two classes confusion matrix}, since Precision and Recall do not consider the True Negative elements, we calculate the binary F1-Score as follows:\medskip

    \begin{equation}
     Precision = \frac{20}{30}=0.66 \  \  \  \  \  \ Recall = \frac{20}{25}=0.80
    \end{equation}\medskip
    
    \begin{equation}
      \textrm{F1-Score}= 2\cdot \left(\frac{0.66 \cdot 0.80}{0.66+0.80}\right)= 0.72
    \end{equation}\medskip

The F1-Score for the binary case takes into account both Precision and Recall. Thanks to these metrics, we can be quite confident that F1-Score will spot weak points of the prediction algorithm, if any of those points exists.\\
In our example a score of 0.72 demonstrates a quite good model ability in predicting the correct class.

\subsection{F1-Score Multi-class case}

When it comes to multi-class cases, F1-Score should involve all the classes. To do so, we require a multi-class measure of Precision and Recall to be inserted into the harmonic mean.
Such metrics may have two different specifications, giving rise to two different metrics: Micro F1-Score and Macro F1-Score \cite{opitz2019macro}.\medskip

\subsubsection{Macro F1-Score}

In order to obtain Macro F1-Score, we need to compute Macro-Precision and Macro-Recall before. They are respectively calculated by taking the average precision for each predicted class and the average recall for each actual class. Hence, the Macro approach considers all the classes as basic elements of the calculation: each class has the same weight in the average, so that there is no distinction between highly and poorly populated classes.
\medskip

\begin{figure}[ht!]
    \centering
    \includegraphics[scale=0.6]{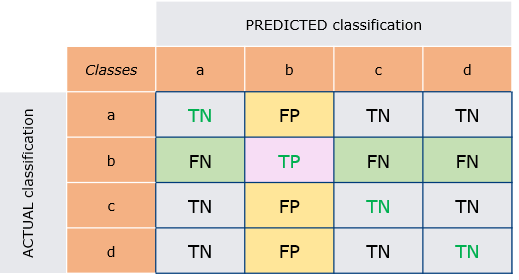}
    \includegraphics[scale=0.52]{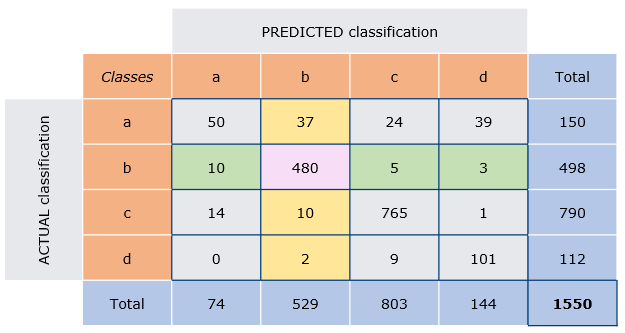}
    \caption{Macro-Precision and Macro-Recall\\ Class $b$ is the reference and we show how to compute its Precision and Recall}
    \label{fig:Macro-Precision & Macro-Recall}
\end{figure}\medskip

For the required computations, we will use the Confusion Matrix focusing on one class at a time and labelling the tiles accordingly. In particular, we consider True Positive (TP) as the only correctly classified units for our class, whereas False Positive (FP) and False Negative (FN) are the wrongly classified elements on the column and the row of the class respectively. True Negative (TN) are all the other tiles, as shown in Figure \ref{fig:Macro-Precision & Macro-Recall} where we are considering the class "b" as reference focus. \\
When we switch from one class to another one, we compute the quantities again and the labels for the Confusion Matrix tiles are changed accordingly. 
\medskip

Precision and Recall for each class are computed using the same formulas of the binary setting and the labelling, as described above. The Formulas \ref{prec_F1} and \ref{rec_F1} represent the two quantities for a generic class $k$.

\begin{equation}\label{prec_F1}
  Precision_k=\frac{TP_k}{TP_k+FP_k}
\end{equation}

\begin{equation}\label{rec_F1}
  Recall_k=\frac{TP_k}{TP_k+FN_k}
\end{equation}
\medskip

Macro Average Precision and Recall are simply computed as the arithmetic mean of the metrics for single classes.

    \begin{equation}
       Macro Average Precision = \frac{ \sum_{k=1}^K Precision_k}{K}
    \end{equation}\medskip
    
    \begin{equation}
       Macro Average Recall = \frac{ \sum_{k=1}^K Recall_k}{K}
    \end{equation}\medskip

Eventually, Macro F1-Score is the harmonic mean of Macro-Precision and Macro-Recall:\medskip

   \begin{equation}
     Macro\ \textrm{F1-Score}=2*(\frac{MacroAveragePrecision*MacroAverageRecall}{MacroAveragePrecision^{-1}+MacroAverageRecall^{-1}})
    \end{equation}\medskip
    
It is possible to derive some intuitions from the equation.\\
Macro-Average methods tend to calculate an overall mean of different measures, because the numerators of Macro Average Precision and Macro Average Recall are composed by values in the range $[0,1]$. There is no link to the class size, because classes with different size are equally weighted at the numerator. This implies that the effect of the biggest classes have the same importance as small ones have. The obtained metric evaluates the algorithm from a class standpoint: high Macro-F1 values indicate that the algorithm has good performance on all the classes, whereas low Macro-F1 values refers to poorly predicted classes.
\medskip

\subsection{Micro F1-Score}

In order to obtain Micro F1-Score, we need to compute Micro-Precision and Micro-Recall before.\\
The idea of Micro-averaging is to consider all the units together, without taking into consideration possible differences between classes. Therefore, the Micro-Average Precision is computed as follows:

\begin{equation}
   Micro\ Average\ Precision = \frac{\sum_{k=1}^K TP_k}{\sum_{k=1}^K Total\;Column_k} = \frac{\sum_{k=1}^K TP_k}{Grand\; Total}
\end{equation}
\medskip

What about the Micro-Average Recall? When we try to evaluate it, we observe the measure is exactly equal to the Micro-Average Precision, in fact summing the two measures rely on the sum of the True Positives, whereas the difference should be in the denominator: we consider the Column Total for the Precision calculation and the Row Total for the Recall calculation. Although, using the units all together ends up in having the Grand Total in both the Formulas.

\begin{equation}
   Micro\ Average\ Recall = \frac{\sum_{k=1}^K TP_k}{\sum_{k=1}^K Total\;Row_k} = \frac{\sum_{k=1}^K TP_k}{Grand\; Total}
\end{equation}\medskip

Long story short, we may see that Micro-Average Precision and Recall are just the same values, therefore the Micro-Average F1-Score is just the same as well (the harmonic mean of two equal values is just the value).

\begin{equation}
   Micro\ Average\ F1 = \frac{\sum_{k=1}^K TP_k}{Grand \;Total}
\end{equation}
\medskip

Taking a look to the formula, we may see that Micro-Average F1-Score is just equal to Accuracy. Hence, pros and cons are shared between the two measures. Both of them give more importance to big classes, because they just consider all the units together. In fact a poor performance on small classes is not so important, since the number of units belonging to those classes is small compared to the dataset size.
\medskip

All in all, we may regard the Macro F1-Score as an average measure of the average precision and average recall of the classes. This measure is calculated at class level, so that each class has the same weight. Small classes are equivalent to big ones and the algorithm performance on them is equally important, regardless of the class size. \\
On the contrary, trying to reverse the concept and build the Micro F1-score, just give us the Accuracy Formula. So that we have a new interpretation of the Accuracy as the average of Precision and Recall above the entire dataset. The Accuracy, as stated above, is calculated at a dataset level and each unit has the same importance. Therefore, the Accuracy gives different importance to different classes, based on their frequency in the dataset.

\subsection{Cross Entropy}

From a theoretical point of view, Cross-Entropy is used to evaluate the similarity between two distribution functions. Consider two generic distributions $p(x)$ and $q(x)$, the Cross-Entropy is given by the formula \ref{continuous}-\ref{discrete}, to respectively suit continuous or discrete $X$ variables.

\begin{equation}\label{continuous}
	H(p,q) = - \int_{D_x} p(x) \log q(x)
\end{equation}

\begin{equation}\label{discrete}
	H(p,q) = - \sum_{D_x} p(x) \log q(x)
\end{equation}

The metric compares the two distributions over the entire domain $D_X$ of the $X$ variable and it only assumes positive values.
 % (in fact $q(x)$ lies in the interval $[0,1]$ and the logarithmic transformation makes it negative the minus in front makes it positive again).\\
In particular, small values of the Cross-Entropy function denote very similar distributions.
\medskip

\begin{figure}[htp!]

\centering
\begin{tabular}{@{}ccc@{}}
\subfloat[Distribution of $Y_i|\mathbf{X}$]{%
  \includegraphics[width=.5\textwidth]{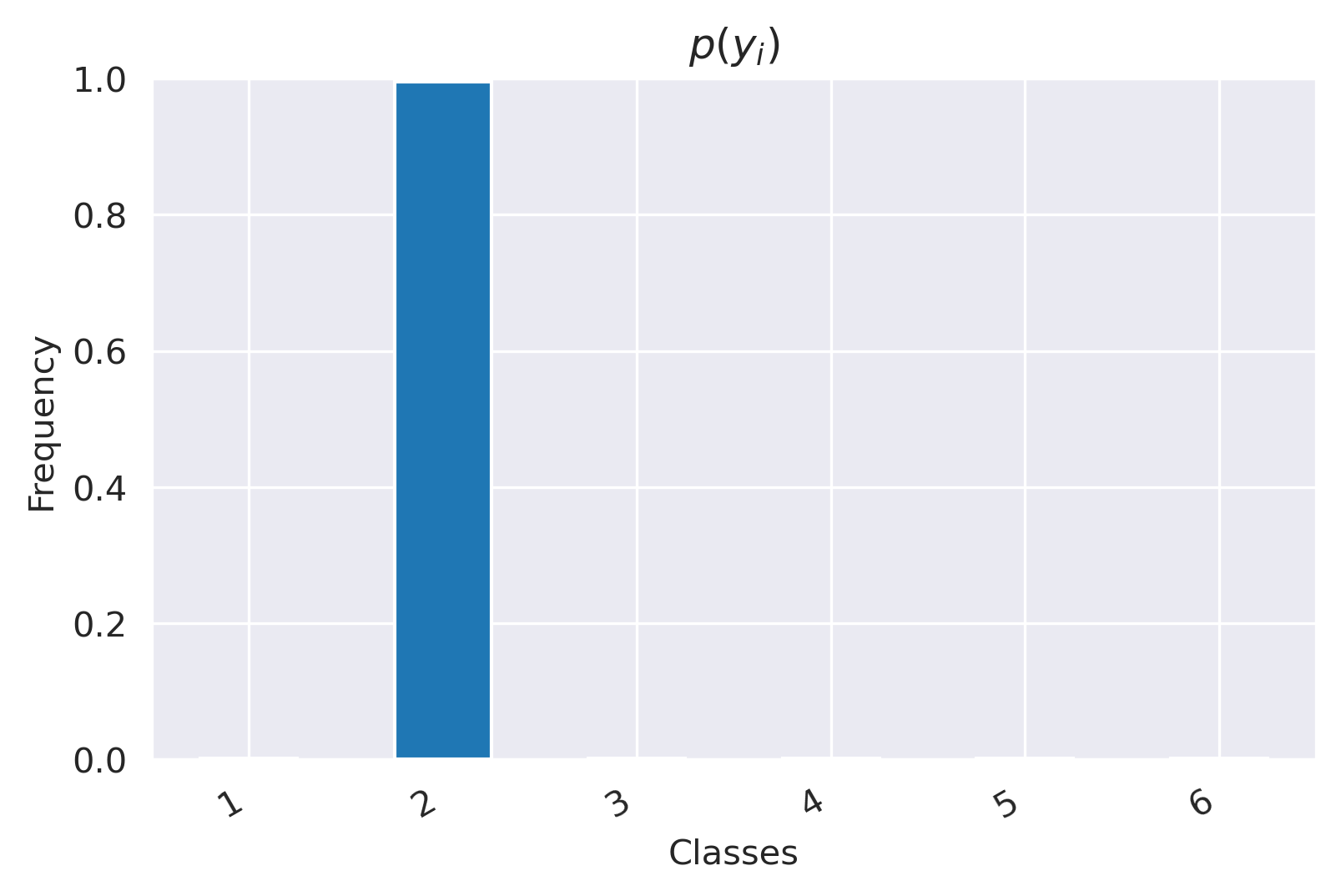}\label{py_i}%
} & 

\subfloat[Distribution of $\hat{Y}_i|\mathbf{X}$]{%
  \includegraphics[width=.5\textwidth]{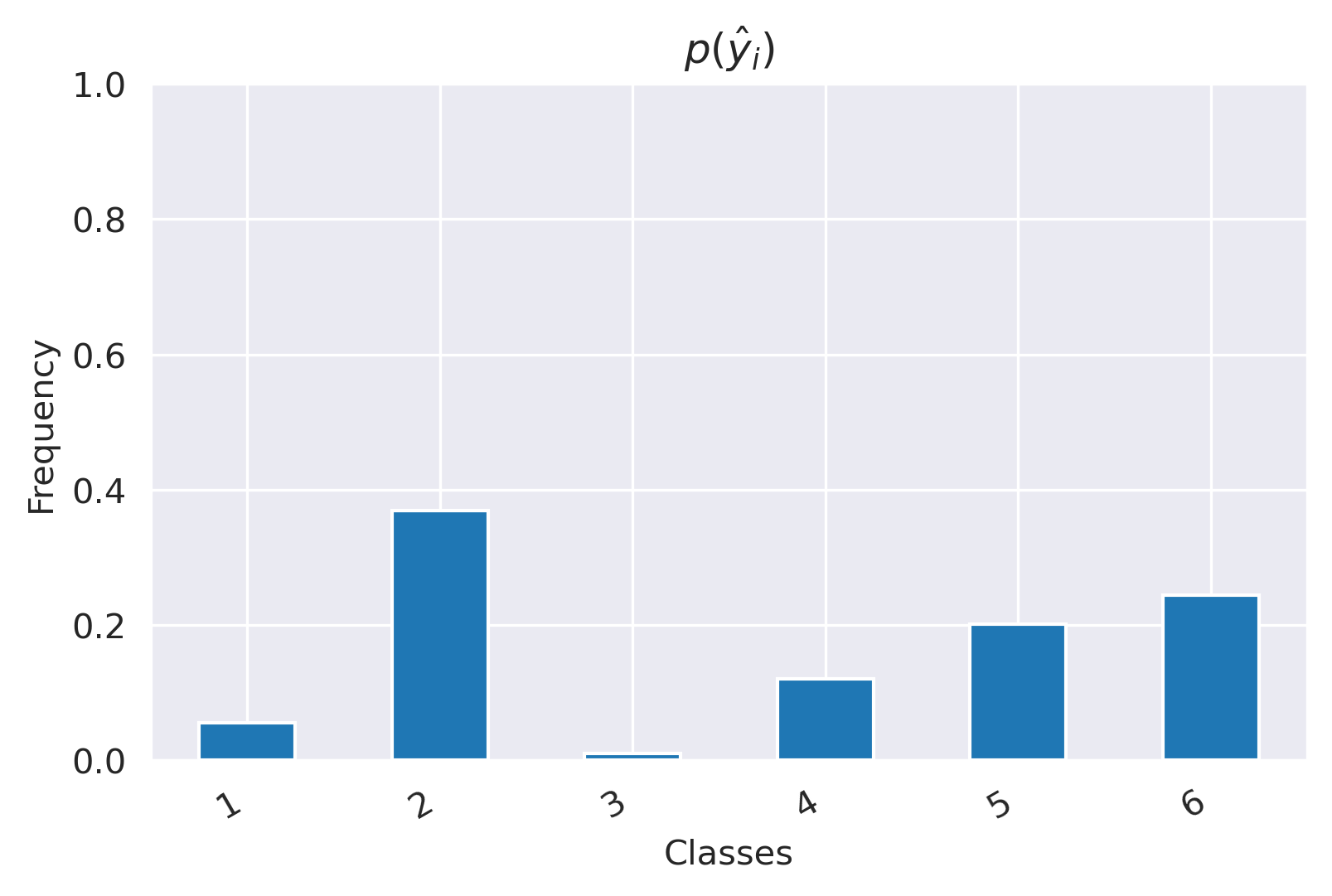}\label{py_i_hat}%
}
\end{tabular}
\caption{}
\label{due_distrib}
\end{figure}

In Multi-class classification, we may regard the response variable $Y$ and the prediction $\hat{Y}$ as two discrete random variables: they assume values in $\{ 1, \cdots, K\}$ and each number represents a different class.\\
Considering the generic $i$-th unit of the dataset: it has specific values $(x_1^{(i)},\cdots,x_m^{(i)})$ of the $\mathbf{X}$ variables and the number $y^{(i)}$ represents the class the unit belongs to. Since we only observe the true class, we consider the unit to have probability equal to 1 for this class and probability equal to 0 for the remaining classes. Doing so, $y^{(i)}$ may be rewritten as a vector of probabilities, as shown in Panel \ref{py_i}. On the other hand, the algorithm prediction itself generates a numeric vector $\hat{y}^{(i)}$, with the probability for the $i$-th unit to belong to each class.
\medskip

$y^{(i)}$ and $\hat{y}^{(i)}$ are generated respectively from the conditioned random variables $Y|\mathbf{X}$ and $\hat{Y}|\mathbf{X}$. The conditioning reflects the fact that we are considering a specific unit, with specific characteristics, namely the unit's values for the $\mathbf{X}$ variables.
\medskip

In the following figures we will regard respectively $p(y_i)$ and $p(\hat{y}_i)$ as the probability distributions of the conditioned variables above. In Figure \ref{due_distrib}, a representation of the two distributions, for a fictitious unit.\\

The Cross-Entropy for the $i$-th unit (Formula \ref{cross-entropy_i}) is calculated between the true distribution $p(y_i)$ and the prediction $p(\hat{y}_i)$.

\begin{equation}\label{cross-entropy_i}
	H(p(y_i),p(\hat{y}_i)) = -\sum_{k=1}^K p(Y_i=k|\mathbf{X_i})  \log{p(\hat{Y_i}=k|\mathbf{X_i})}
\end{equation}

Each class is considered in the formula above, however the quantity $p(Y_i=k|\mathbf{X_i})$ is 0 for all the classes except the true one, making all the terms but one disappear. Cross-Entropy exploits only the value of $p(\hat{Y_i}=k|\mathbf{X_i})$ for the $k$ value representing the true class. 
\medskip

Eventually we consider the average of the Cross-Entropy values for the single units, to obtain a measure of agreement on the entire dataset (Formula \ref{cross-entropy_dataset}).

\begin{equation}\label{cross-entropy_dataset}
	H(p(y),p(\hat{y})) = -\sum_i^N\sum_{k=1}^K p(Y_i=k|\mathbf{X_i})  \log{p(\hat{Y_i}=k|\mathbf{X_i})}
\end{equation}\medskip

It is worth noting that the technique does not rely on the Confusion Matrix, instead it employs directly the variables $Y$ and $\hat{Y}$. Therefore, Cross-Entropy does not evaluate the goodness of the classification rule (the rule which translates the probabilities into the predicted class).
\medskip

\begin{figure}[htp!]

\centering
\begin{tabular}{@{}ccc@{}}
\subfloat[Distribution of $\hat{Y}^{(1}_{i}|\mathbf{X}$]{%
  \includegraphics[width=.5\textwidth]{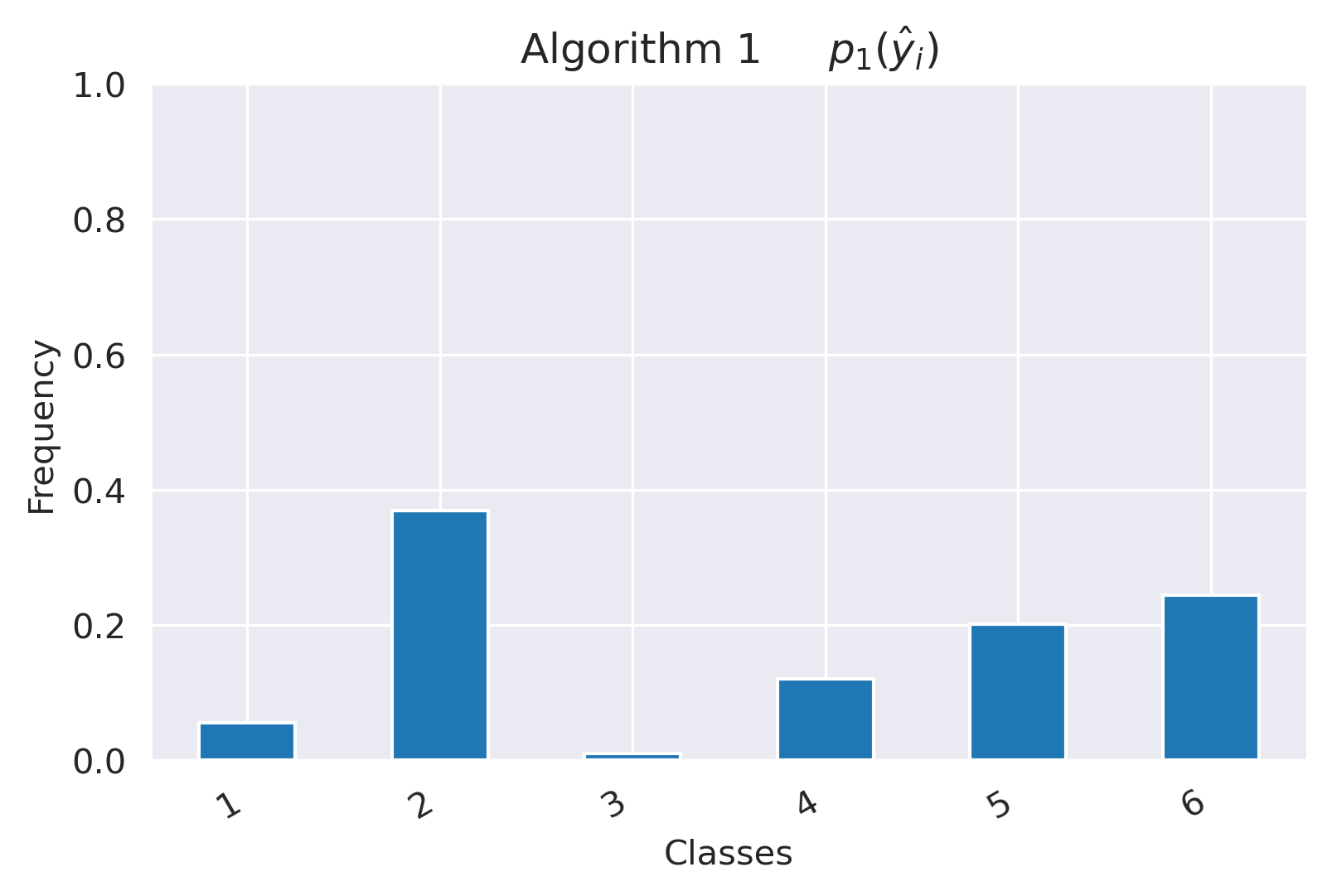}\label{py_i_hat1}%
} & 

\subfloat[Distribution of $\hat{Y}^{(2}_i|\mathbf{X}$]{%
  \includegraphics[width=.5\textwidth]{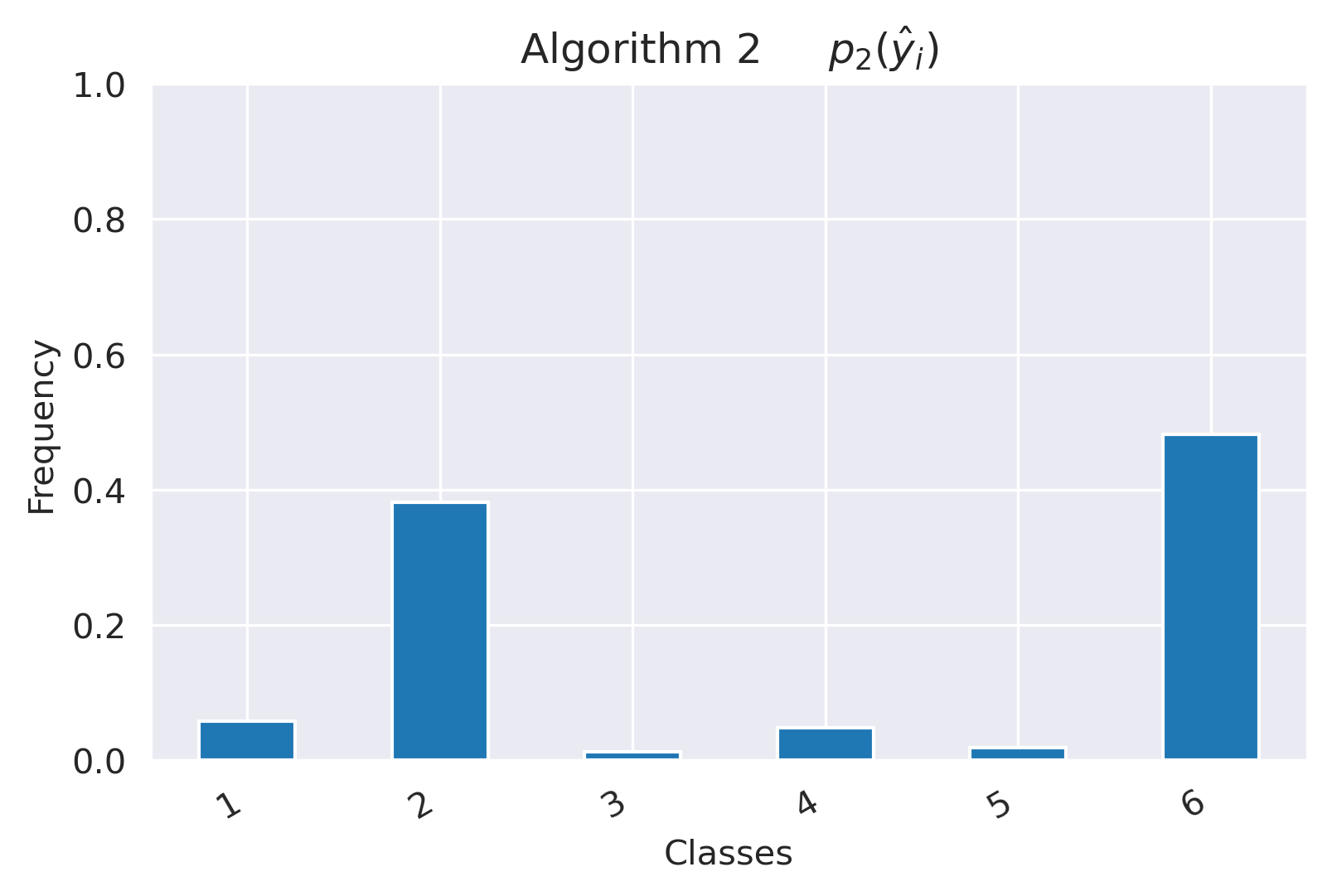}\label{py_i_hat2}%
}
\end{tabular}
\caption{}
\label{drawback_CE}
\end{figure}

From a practical perspective, Cross-Entropy is widely employed thanks to its fast calculation. Although, it takes into account only the true class probability $p(\hat{y}_i=k)$ without caring about the probability mass distribution among the remaining classes. \\
This may have some drawbacks, as shown in Figure \ref{drawback_CE}: the $i$-th unit gets predicted by two different algorithms, obtaining two distinct distributions. The true label is $y_i=2$, referring to the same unit of Figure \ref{due_distrib}.\\

The two algorithms have the same prediction for class 2, i.e. $0.4$, but, substantially, they have different performance on the aggregate perspective: in Panel \ref{py_i_hat1} the highest probability class is 2, for \ref{py_i_hat2} it is 6. Depending on the classification rule, the two algorithms are likely to assign different predicted classes to the unit.
\medskip

Regardless of such differences, Cross-Entropy assigns the same value to each algorithm, since $p(\hat{y}_i=2)$ is the same. In our example, Panel \ref{py_i_hat1} achieves right predictions and \ref{py_i_hat2} a wrong one, without being reported by the Cross-Entropy.

\section{Independence between two Random Discrete Variables}
The last two metrics in this paper were built starting from the confusion matrix and relying on two different statistical concepts. Matthews Correlation Coefficient takes advantage of the Phi-Coefficient \cite{MATTHEWS1975442}, while Cohen's Kappa Score relates to the probabilistic concept of dependence between two random variables. It will be demonstrated that both the scores come up with a measure of how much the model's predictions are dependent on the ground truth classification of a given dataset. Moreover, both the metrics take into account the True Negative (TN) values in the binary case, so they may be preferable to F1-Score when the aim is to assessing the performance of a binary classifier. 

\subsection{Mattheus Correlation Coefficient}
Brian W. Mattheus developed the Mattheus Correlation Coefficient (MCC) in 1975, exploiting Karl Pearson's Phi-Coefficient in order to compare different chemical structures. Only in the 2000s MCC became a widely employed metric to test the performance of Machine Learning techniques with some extensions to the multi-class case \cite{Chicco2020}.
\medskip

MCC has a range of $[-1,1]$. Values close to 1 indicate very good prediction, in fact there is a strong positive correlation between the Prediction and the True Labels. Strong correlation implies that the two variables strongly agree, therefore the predicted values will be very similar to the Actual Classification. On the contrary, when MCC is equal to 0, there is no correlation between our variables: the classifier is randomly assigning the units to the classes without any link to their true class value. \cite{10.1371/journal.pone.0177678}.\\
MCC may also be negative, in this case the relation between true and predicted classes is of an inverse type. Even if this is an highly undesirable situation, this often happens because of setting errors in the modelling: strong inverse correlation means that the model learnt how to classify the data but it systematically switches all the labels. But it is also possible to solve the problem by fixing the implementation errors.\medskip

\subsubsection{Mattheus Correlation Coefficient for binary classification}
MCC could be seen as the Phi-Coefficient applied to binary classification problems: as described above, we consider the "Predicted" classification and "Actual" classification as two discrete random variables and we evaluate their association.
\medskip

\begin{figure}[ht!]
  \centering
  \includegraphics[scale=0.70]{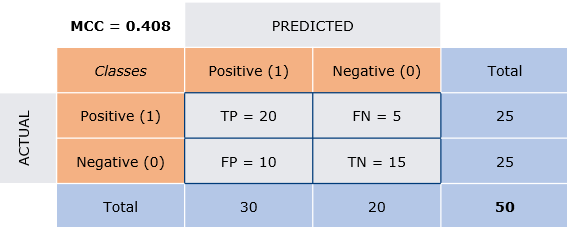}
  \caption{Predicted and Actual Random Variables}
  \label{fig:Predicted and Actual Random Variables}
\end{figure}

\begin{equation}\label{mcc_binary}
       MCC = \frac{TP\times TN - FP\times FN}{\sqrt{(TP+FN)(TP+FP)(TN+FN)(TN+FP)}}
\end{equation}

In Formula \ref{mcc_binary}, we notice that MCC takes into account all the confusion matrix cells. In particular, the numerator consists of two products including all the four inner cells of the confusion matrix in Figure \ref{fig:Predicted and Actual Random Variables}, while the denominator consists of the four outer cells (row totals and column totals).\medskip

A practical demonstration of the concept that there is an effective support regarding the equivalence of MCC and Phi-coefficient in the binary case is given by \cite{Nechushtan2020}.

% , and it coincides with the following:\medskip

% $$
% \quad \rho=\frac{E(x y)-E(x) E(y)}{\sigma(x) \sigma(y)}=\frac{E(x y)-E(x) E(y)}{\sqrt{\left(E\left(x^{2}\right)-E^{2}(x)\right)\left(E\left(y^{2}\right)-E^{2}(y)\right)}}
% a=p(x=1)
% $$

% $$
% \begin{array}{l}
% a=p(y=1) \\
% c=p(x=1 \wedge y=1) \\
% m= \text{Number of examples} \\
% E(x)=E\left(x^{2}\right)=a \\
% E(y)=E\left(y^{2}\right)=b \\
% E(x y)=c
% \end{array}
% $$

% $$
% \rho=\frac{c-a b}{\sqrt{\left(a-a^{2}\right)\left(b-b^{2}\right)}}=\frac{c-a b}{\sqrt{a(1-a) b(1-b)}}=\frac{c-a b}{\sqrt{a b(1-a)(1-b)}}
% $$

% $$
% \begin{array}{l}
% a=(T P+F P) / m \\
% b=(T P+F N) / m \\
% c=(T P) / m \\
% \end{array}
% $$

% $$
% \begin{aligned}
% \rho=& \frac{T P / m-(T P+F P)(T P+F N) / m^{2}}{\sqrt{(T P+F P)(T P+F N) / m^{2}(F N+T N)(F P+T N) / m^{2}}} \\
% \end{aligned}
% $$

% $$
% \begin{aligned}
% &=\frac{T P \cdot m-(T P+F P)(T P+F N)}{\sqrt{(T P+F P)(T P+F N)(F N+T N)(F P+T N)}} \\
% \end{aligned}
% $$

% $$
% \begin{aligned}
% =& \frac{T P \cdot(T P+F P+T N+F N)-(T P+F P)(T P+F N)}{\sqrt{(T P+F P)(T P+F N)(F N+T N)(F P+T N)}}=\\
% \end{aligned}
% $$
% $$
% \begin{aligned}
% =&\frac{(T P \cdot T P)+\{T P \cdot F P\}+T P \cdot T N+[T P \cdot F N]-(T P \cdot T P)-[T P \cdot F N]-\{F P \cdot T P\}-F P \cdot F N}{\sqrt{(T P+F P)(T P+F N)(F N+T N)(F P+T N)}} \\
% \end{aligned}
% $$
% $$
% \begin{aligned}
% =& \frac{T P \cdot T N-F P \cdot F N}{\sqrt{(T P+F P)(T P+F N)(F N+T N)(F P+T N)}}
% \end{aligned}
% $$\\\\
% \medskip

\subsubsection{Mattheus Correlation Coefficient for Multi-class Classification}

Some changes happen when it comes to multi-class classification: the True and the Predicted class distributions are no longer binary and a higher number of classes has been taken into account. In this case numerator and denominator take a different shape compared to the binary case and this can partially help to find more stable results inside the range [-1; +1] of MCC.\medskip

In the multi-class case, the Matthews correlation coefficient can be defined in terms of a confusion matrix \textit{C} for \textit{K} classes.\medskip

\begin{figure}[ht!]
  \centering
  \includegraphics[scale=0.70]{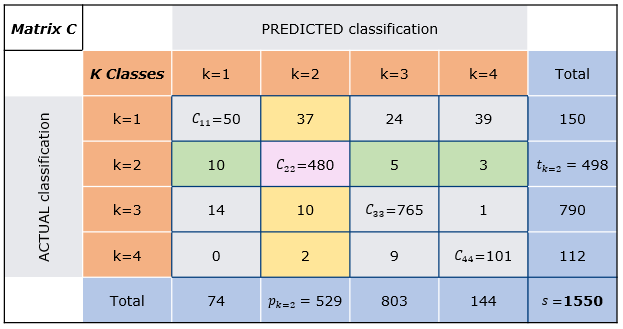}
  \caption{Multi-class Confusion Matrix $C$}
  \label{Multi-class Confusion Matrix}
\end{figure}\medskip

\begin{equation}
       MCC = \frac{c\times s - \sum_{k}^K p_k \times t_k}{\sqrt{(s^2 - \sum_{k}^K p_k^2)(s^2 - \sum_{k}^K t_k^2)}}
\end{equation}
\medskip

To simplify the definition, it is necessary to consider the following intermediate variables \cite{GORODKIN2004367}:\medskip

\begin{itemize}
    \item $c=\sum_{k}^K C_{kk}$ the total number of elements correctly predicted
    \item $s=\sum_{i}^K \sum_{j}^K C_{ij}$ the total number of elements 
    \item $p_k=\sum_{i}^K C_{ki}$ the number of times that class $k$ was predicted (column total)
    \item $t_k=\sum_{i}^K C_{ik}$ the number of times that class $k$ truly occurred (row total)
\end{itemize}
\medskip
\medskip

The setting of this formula has raised some intuitions.\\
In the multi-class case MCC seems to depend on correctly classified elements, because the total number of elements correctly predicted are multiplied by the total number of elements at the numerator and the weight of this product is more powerful than the sum $\sum_{k}^K p_k \times t_k$. This sum includes also the elements wrongly classified by the model and covers multiplicative entities that are weaker than the product $c\times s$. \medskip

Regarding the denominator, it is employed to rescale the fraction in the interval $[-1,+1]$, in fact it corresponds to the maximum absolute value the numerator may assume. The first term in the denominator entirely depends on the Predicted classification, whereas the second one depends on the True classes, which can be considered as property of the dataset since they do not change when we apply different models on the same dataset.\medskip

The weakness of MCC involves its lower limits. There is no fixed minimum value and it changes every time between -1 and 0 depending on the number and on the Actual distribution of the classes in the initial dataset \cite{GORODKIN2004367}.\medskip

\subsubsection{Pros and Cons of MCC}

Among the Advantages of this technique, we can see that MCC includes all the entries of the confusion matrix both at the numerator and the denominator. This means that MCC is generally regarded as a balanced measure which can be used in binary classification even if the classes are very different in size \cite{Chicco2020}.
\medskip

\begin{figure}[ht!]
          \centering
          \includegraphics[scale=0.70]{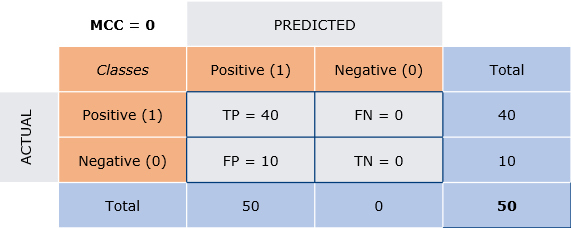}
          \caption{Total Unbalanced Prediction}
          \label{fig:Total Unbalanced Prediction}
     \end{figure}

Moreover, MCC is a good indicator of total unbalanced prediction models. We have shown this topic in Figure \ref{fig:Total Unbalanced Prediction}, where the model assigns all the elements to only one class and the value of MCC falls to 0, even if the Accuracy achieves a great value (0.80) and the Recall for the first class assumes the highest value (1).

     \medskip

However, the weaknesses of MCC could be found in some extreme cases and they are mainly related to its construction. If there are unbalanced results in the model's prediction, the final value of MCC shows very wide fluctuations inside its range of [-1; +1] during the training period of the model \cite{doi:10.1002/minf.201700127}.

%%% aggiungere due paper

\subsection{Cohen's Kappa}
Cohen’s Kappa builds on the idea of measuring the concordance between the Predicted and the True Labels, which are regarded as two random categorical variables \cite{Ranganathan2017}. It is possible to compare two categorical variables building the confusion matrix and calculating the marginal rows and the marginal columns distributions. \medskip

\begin{figure}[ht!]
  \centering
  \includegraphics[scale=0.70]{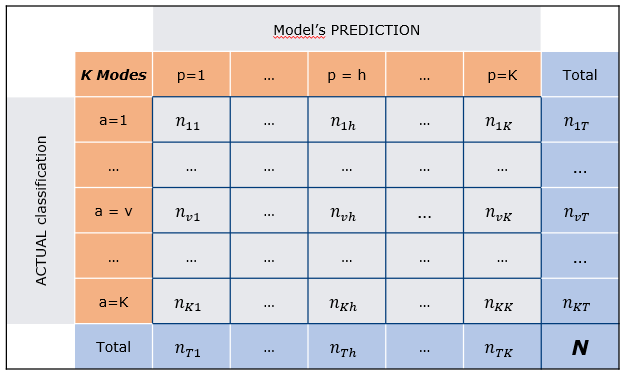}
  \caption{Confusion matrix for two General Categorical Distributions}
  \label{Confusion matrix for two General Categorical Distributions}
\end{figure}\medskip

In particular two distributions of the same character are independent if they assume the same relative frequencies at the same character model.

\begin{equation}
       \frac{n_{vh}}{n_{Th}} = \frac{n_{vT}}{N}
\end{equation}

Also, two characters (i.e. model's Prediction \& Actual classification) are independent variables in distribution if this relationship is true:

\begin{equation}
      {n_{vh}}^*= \frac{n_{Th} \times n_{vT}}{N} 
\end{equation}

And ${n_{vh}}*$ stands for a relative frequency that we expect to find if two categorical distributions are independent.\medskip

Given this definition of independence between categorical variables, we can start dealing with Cohen's Kappa indicators as rating values of the dependence (or independence) between the model's Prediction and the Actual classification.\medskip

The marginal columns distribution can be regarded as the distribution of the Predicted values (how many elements are predicted in each possible class), while the Marginal rows represent the distribution of the True classes.
\medskip

Moreover, we will see in this chapter why Cohen's Kappa could be also useful in evaluating the performance of two different models when they are applied on two different databases and it allows to make a comparison between them.

\subsubsection{Cohen's Kappa for binary classification}
Starting from a simple confusion matrix:
\medskip

\begin{figure}[ht!]
 \centering
 \includegraphics[scale=0.70]{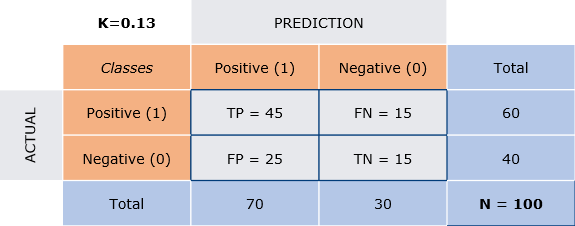}
 \caption{Confusion Matrix for binary Cohen's Kappa}
 \label{fig:Confusion Matrix for binary Cohen's Kappa}
\end{figure}\medskip

Cohen (1960) evaluated the classification of two raters (i.e. model's Prediction \& Actual distribution) in order to find a measure of agreement between them, or rather, he looked for a statistic giving the degree of concordance between two (or more) sets of measurements. He calculated the inter-observer agreement taking into account the expected agreement by chance as follows
\cite{Ranganathan2017}:\medskip

\begin{equation}
       K = \frac{P_o-P_e}{1-P_e}
\end{equation}\medskip

Where
\begin{itemize}
    \item $P_o$ is the proportion of observed agreement, in other words it is the Accuracy achieved by the model
    \item $P_e$ is the Expected Accuracy, i.e. the level of Accuracy we expect to obtain by chance. If we have a model that classifies the units in random classes, preserving just the distribution of the predicted classes, its Accuracy should be close to $P_e$
    \item $1-P_e$ is essentially the difference between the maximum value and the minimum value of the numerator, in this way we are re-scaling the final index between -1 and +1.  
\end{itemize}
\medskip

\begin{equation}
       P_e = P_{Positive} + P_{Negative}
\end{equation}
\medskip
\begin{equation}
       P_{Positive}= \frac{TP+FN}{N}\times  \frac{TP+FP}{N}
\end{equation}
\medskip
\begin{equation}
      P_{Negative} = \frac{TN+FP}{N}\times  \frac{TN+FN}{N}
\end{equation}
\medskip

The K statistic can take values from $-1$ to $+1$ and is interpreted somewhat arbitrarily as follows: 0 is the agreement equivalent to chance, from 0.10 to 0.20 is a slight agreement, from 0.21 to 0.40 is a fair agreement, from 0.41  to 0.60 is a moderate agreement, from 0.61 to 0.80 is a substantial agreement, from 0.81 to 0.99 is a near perfect agreement and 1.00 is a perfect agreement. Negative values indicate that the observed agreement is worse than what would be expected by chance. An alternative interpretation is offered by \cite{Ranganathan2017} saying that kappa values below 0.60 indicate a significant level of disagreement.\medskip\medskip

\begin{itemize}
    \item \textbf{Insights}
\end{itemize}
\medskip

Given the similarity of the last operations to the concept of independence between two events,
\medskip
\begin{equation}
      P_{Positive} = P(Prediction_{1} \cap Actual_{1})
\end{equation}
\begin{equation}
      P_{Positive}= P(Prediction_{1}) \times P(Actual_{1})
\end{equation}
\begin{equation}
      P_{Positive} = \frac{45+15}{100}\times  \frac{45+25}{100} = 0.42
\end{equation}\medskip

we have noticed that the Expected Accuracy $P_e $ plays the main role in the Cohen’s Kappa Score because it brings with it two components of independence ($P_{Positive}$ and $P_{Negatives}$) which are subtracted from the observed agreement $P_o$.\medskip

It is important to remove the Expected Accuracy (the random agreement component for Cohen and the two independent components for us) from the Accuracy for two reasons: the Expected Accuracy is related to a classifier that assigns units to classes completely at random, it is important to find a model's Prediction that is as dependent as possible to the Actual distribution. So Cohen's Kappa results to be a measure of how much the model's prediction is dependent on the Actual distribution, with the aim to identify the best learning algorithm of classification.
\medskip

These are the basic intuitions on Cohen's Kappa score and they have to be supported by demonstrations:\medskip

\begin{equation}\label{expected_accuracy}
       P_e = P_{Positive} + P_{Negative}
\end{equation}
\medskip

\begin{equation}
       P_e = \frac{TP+FN}{N}\times  \frac{TP+FP}{N} + \frac{TN+FP}{N}\times  \frac{TN+FN}{N}
\end{equation}
\medskip

\begin{equation}
      if\ {n_{vh}}^*= \frac{n_{Th} \times n_{vT}}{N} \ 
      then\ {TP}^* = \frac{(45+15) \times (45+25)}{100}
\end{equation}
\medskip

\begin{equation}
      and\ P_{Positive}= {TP}^* \times \frac{1}{100} \ and\
       P_{Negative}= {TN}^* \times \frac{1}{100}
\end{equation}
\medskip

Coming back to the formula \ref{expected_accuracy}: \medskip

\begin{equation}
       P_e = P_{Positive} + P_{Negative} = \frac{{TP}^*}{100} + \frac{{TN}^*}{100} = \frac{{TP}^* + {TN}^*}{100}
\end{equation}
\medskip

If two random and categorical variables are independent they should have this Accuracy $\frac{{TP}^* + {TN}^*}{100}$.
But, since we want that the Predicted and Actual distribution to be as dependent as possible, Cohen's Kappa score directly subtracts this previous Accuracy from the observed agreement at the numerator of the formula. In this way, we have obtained an Accuracy value related only to the goodness of the model and we have already deleted the part ascribed to chance (the Expected Accuracy).\medskip

Just as a reminder, two dependent variables are also correlated and identified by reciprocal agreement. In our case a high correlation is observed when the model's Prediction assigns a unit to one class, and the same unit has been also assigned to the same class by the Actual classification.

\subsubsection{Cohen's Kappa for multi-class cases}
In the multi-class case, the calculation of Cohen's Kappa Score changes its structure and it becomes more similar to Mattheus Correlation Coefficient \cite{dataminingmethods}.

Referring to Multi-class Confusion Matrix \textit{C} in Figure \ref{Multi-class Confusion Matrix}:\medskip

\begin{equation}\label{K_multi}
       K = \frac{c\times s - \sum_{k}^K p_k \times t_k}{s^2 - \sum_{k}^K p_k\times t_k}
\end{equation}
\medskip

Where:
\begin{itemize}
    \item $c=\sum_{k}^K C_{kk}$ the total number of elements correctly predicted
    \item $s=\sum_{i}^K \sum_{j}^K C_{ij}$ the total number of elements 
    \item $p_k=\sum_{i}^K C_{ki}$ the number of times that class $k$ was predicted (column total)
    \item $t_k=\sum_{i}^K C_{ik}$ the number of times that class $k$ truly occurs (row total)
\end{itemize}
\medskip
MCC and Cohen's Kappa coincides in the multi-class cases apart from the denominator that is slightly lower in Cohen's Kappa score justifying slightly higher final scores. However some evidences of the binary case still holds: when $K$ is equal to 0 the model's Prediction is totally independent from the Actual classification and if $K$ is equal to 1 the model's Prediction is totally dependent from the Actual classification. Instead $K$ is negative when the agreement between the algorithm and the true labels distribution is worse than the random agreement, so that there is no accordance between the model's Prediction and the Actual classification.\medskip

As before, the advantage of Cohen's Kappa score must be sought through the measure of Expected Accuracy as an intrinsic characteristic of a given dataset. In the multi-class case the Expected Accuracy assumes the shape of the sum applied on the row and column totals multiplication for each class $k$ (\ref{K_multi}).\\

\subsubsection{Useful Applications}

\begin{figure}[ht!]
 \centering
 \includegraphics[scale=0.70]{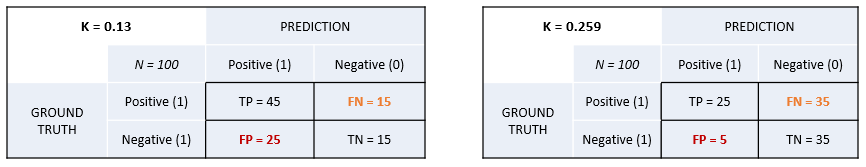}
 \caption{Cohen's Kappa Matrix for comparison}
 \label{Cohen's Kappa Matrix for comparison}
\end{figure}

Cohen's Kappa finds useful applications in many classification problems.\\ Firstly it allows the joint comparison of two models for which it has registered the same accuracy, but different values of Cohen's Kappa. Figure \ref{Cohen's Kappa Matrix for comparison} is a simplified binary example, where $K$ increases more the errors are unbalanced towards one class. This is true also for multi-class settings. \medskip

Secondly, the Expected Accuracy re-scales the score and represents the intrinsic characteristics of a given dataset. \\
We consider both the number of classes and the fact to be balanced or unbalanced towards a group of classes as the two main representative characteristics of a dataset. Subtracting the Expected Accuracy we are also removing the intrinsic dissimilarities of different datasets and we are making two different classification problems comparable. As a result, $K$ can compare the performances of two different model on two different cases.

\section{Conclusions}
The aim of this essay is an in-depth analysis of different choices to evaluate the performance of different classification algorithms on multi-class datasets.

As the most famous classification performance indicator, the \textbf{Accuracy} returns an overall measure of how much the model is correctly predicting the classification of a single individual above the entire set of data. It is an average measure which is suitable for balanced datasets because it does not consider the class distribution.

As a simple arithmetic mean of Recalls, the \textbf{Balanced Accuracy} gives the same weight to each class and its insensibility to class distribution helps to spot possible predictive problems also for rare and  under-represented classes. 

As weighted average of Recall, the \textbf{Balanced Accuracy Weighted} keeps track of the importance of each class thanks to the frequency. In this case, large and small classes have a proportional effect on the result in relation to their size and the metric can be applied during the training phase of the algorithm on a wide number of classes.

As harmonic mean of Macro Precision and Macro Recall, \textbf{Macro-Average} methods tend to calculate an overall mean of different measures without taking into account the class size. The effect of the biggest classes is shifted by the smallest ones which have the same weight.

Regarding \textbf{Micro F1-Score}, it is possible to show that the harmonic mean of Micro Precision and Micro Recall just boils to the Accuracy formula, giving a new interpretation of it.

As average of \textbf{Cross Entropy} for each unit in a dataset, it is a measure of agreement between two probability distributions (predicted and true classification). Cross Entropy is detached from the confusion matrix and it is widely employed thanks to his fast calculation. Although it just takes into account the prediction probability of the right class, without considering how the probability distribution behaves on the other classes, this may cause issues especially when a unit is misclassified.

It may be considered as the successor of Karl Pearson's Phi-Coefficient, the \textbf{Mattheus Correlation Coefficient} expresses the degree of correlation between two categorical random variables (predicted and true classification). Its result covers the range [-1; +1] pointing out different model behaviors during the training phase of the algorithm.

Its value represents the dependence between the predicted and the true classification, \textbf{Cohen's Kappa} exploits the Expected Accuracy, namely a measure representing the dependence obtained by chance between the predicted and the true classification measure, to delete any intrinsic characteristic of the dataset. This allows for the comparison between different models applied on different samples of data. 

\section*{Funding}
We acknowledge financial support by CRIF S.p.A. and Università degli Studi di Bologna.

\printbibliography

\end{document}